\documentclass{article}

\usepackage{arxiv}

\usepackage[utf8]{inputenc} 
\usepackage[T1]{fontenc}    
\usepackage{hyperref}       
\usepackage{url}            
\usepackage{booktabs}       
\usepackage{array}
\usepackage{multirow}
\usepackage{amsfonts}       
\usepackage{nicefrac}       
\usepackage{microtype}      
\usepackage{lipsum}
\usepackage{graphicx}
\graphicspath{ {./images/} }
\usepackage{booktabs}
\usepackage{xcolor}
\usepackage{svg}
\usepackage{rotating}
\usepackage{subcaption}
\usepackage{amsmath, amssymb, algorithm, algpseudocode}
\usepackage{geometry}

\newcolumntype{L}[1]{>{\raggedright\arraybackslash}p{#1}}

\usepackage[numbers]{natbib} 

\usepackage[acronym]{glossaries} 

\title{A Hybrid Inductive-Transductive Network for Traffic Flow Imputation on Unsampled Locations}

\author{
 Mohammadmahdi Rahimiasl \\
  University of Antwerp - imec, IDLab - Faculty of Applied Engineering\\\\
  Sint-Pietersvliet 7, 2000 Antwerpen\\\\
  \texttt{Mohammadmahdi.Rahimiasl@uantwerpen.be} \\
   \And
 Ynte Vanderhoydonc \\
  University of Antwerp\\\\
  Middelheimlaan 1, 2020 Antwerpen\\\\
  \texttt{Ynte.Vanderhoydonc@uantwerpen.be} \\
  \And
 Siegfried Mercelis \\
  University of Antwerp - imec, IDLab - Faculty of Applied Engineering\\\\
  Sint-Pietersvliet 7, 2000 Antwerpen\\\\
  \texttt{Siegfried.Mercelis@uantwerpen.be} \\
}

\begin{document}
\maketitle
\newacronym{fcd}{FCD}{Floating Car Data}
\newacronym{ml}{ML}{Machine Learning}
\newacronym{dl}{DL}{Deep Learning}
\newacronym{gnn}{GNN}{Graph Neural Networks}
\newacronym{lstm}{LSTM}{Long Short-Term Memory}
\newacronym{gru}{GRU}{Gated Recurrent Unit}
\newacronym{rnn}{RNN}{Recurrent Neural Network}
\newacronym{mae}{MAE}{Mean Absolute Error}
\newacronym{aadt}{AADT}{Annual Average Daily Traffic}
\newacronym{rf}{RF}{Random Forest}
\newacronym{svm}{SVM}{Support Vector Machine}
\newacronym{ann}{ANN}{Artificial Neural Network}
\newacronym{gcn}{GCN}{Graph Convolutional Network}
\newacronym{tcn}{TCN}{Temporal Convolutional Network}
\newacronym{dgcn}{DGCN}{Diffusion Graph Convolution Networks}
\newacronym{selu}{SELU}{Scaled Exponential Linear Unit}
\newacronym{relu}{ReLU}{Rectified Linear Unit}
\newacronym{mape}{MAPE}{Mean Absolute Percentage Error}
\newacronym{smape}{SMAPE}{Symmetric Mean Absolute Percentage Error}
\newacronym{rmse}{RMSE}{Root Mean Squared Error}
\newacronym{pce}{PCE}{Passenger Car Equivalent}
\newacronym{tpe}{TPE}{Tree-structured Parzen Estimator}
\newacronym{knn}{KNN}{K-Nearest Neighbour}
\newacronym{gpr}{GPR}{Gaussian Process Regression}
\newacronym{mow}{MOW}{Meten-in-Vlaanderen: minuutwaarden verkeersmetingen}
\newacronym{osm}{OSM}{OpenStreetMap}
\newacronym{cnn}{CNN}{Convolutional Neural Network}
\newacronym{nlp}{NLP}{Natural Language Processing}
\newacronym{iqr}{IQR}{Interquartile Range}
\newacronym{svr}{SVR}{Support Vector Regression}
\newacronym{gbdt}{GBDT}{Gradient Boosting Decision Tree}
\newacronym{var}{VAR}{Vector Autoregressive}
\newacronym{arima}{ARIMA}{Autoregressive Integrated Moving Average}
\newacronym{pbf}{PBF}{Protocolbuffer Binary Format}
\newacronym{film}{FiLM}{Feature-wise Linear Modulation}
\newacronym{sta}{STA}{Static Traffic Assignment}
\newacronym{bpr}{BPR}{Bureau of Public Roads} 

\begin{abstract}
Accurately imputing \textbf{traffic flow} at unsensed locations is difficult: loop detectors provide precise but sparse measurements, speed from probe vehicles is widely available yet only weakly correlated with flow, and nearby links often exhibit strong \textbf{heterophily} in the scale of traffic flow (e.g., ramps vs.\ mainline), which breaks standard GNN assumptions. We propose \textbf{HINT}, a \textbf{H}ybrid \textbf{IN}ductive-\textbf{T}ransductive Network, and an \textbf{INDU-TRANSDUCTIVE} training strategy that treats speed as a transductive, network-wide signal while learning flow inductively to generalize to unseen locations. HINT couples (i) an inductive spatial transformer that learns similarity-driven, long-range interactions from node features with (ii) a diffusion GCN conditioned by FiLM on rich static context (OSM-derived attributes and traffic simulation), and (iii) a node-wise calibration layer that corrects scale biases per segment. Training uses masked reconstruction with epoch-wise node sampling, hard-node mining to emphasize difficult sensors, and noise injection on visible flows to prevent identity mapping, while graph structure is built from driving distances.

Across three real-world datasets, MOW (Antwerp, Belgium), UTD19-Torino, and UTD19-Essen, HINT consistently surpasses state-of-the-art inductive baselines. Relative to KITS, HINT reduces MAE on MOW by \textbf{$\approx42$\%} with basic simulation and \textbf{$\approx50$\%} with calibrated simulation; on Torino by \textbf{$\approx22$\%}, and on Essen by \textbf{$\approx12$\%}. Even without simulation, HINT remains superior on MOW and Torino, while simulation is crucial on Essen. These results show that combining inductive flow imputation with transductive speed, traffic simulations and external geospatial improves accuracy for the task described above.
\end{abstract}

\section{Introduction}

Traffic flow is a measure defined as vehicle counts passing through specific road segments over time. Reliable traffic flow data is essential for intelligent transportation planning, effective congestion management, and optimization of traffic control strategies. However, obtaining comprehensive and accurate traffic flow data remains challenging due to practical limitations associated with existing data collection methods.

Traditional devices, notably inductive loop detectors deliver highly accurate data independent of environmental factors such as weather or visibility. Yet, these methods suffer from high installation and maintenance costs that leads to limited spatial coverage \citep{liang2019urbanfm, xu2023kits}. In contrast, \gls{fcd} services primarily derived from navigation applications offer extensive spatial coverage but they mainly capture speed rather than flow. \gls{fcd} struggles to provide reliable flow estimates because its coverage heavily depends on the number of active app users which varies by region and over time. This issue is particularly pronounced in regions with low application penetration rates.

One widely used conceptual framework in traffic analysis is the speed-flow or the fundamental diagram, which illustrates the complex and often nonlinear relationship between average speed and traffic flow. Historically, such diagrams have been modeled by foundational approaches like Greenshields, Pipes, and Van Aerde \citep{rakha2002comparison}, each attempted to capture macroscopic and microscopic flow characteristics. However, in many real-world scenarios, higher flow does not produce significantly lower speeds until near-congested conditions that complicates direct inferences of flow from speed data \citep{neumann2013dynamic, bulteau2013traffic}.

Motivated by these challenges, several studies have investigated traffic flow estimation by exploring alternative data sources and modeling techniques. \citet{bulteau2013traffic} developed a methodology leveraging higher order speed statistics (e.g., inter-lane speed variance) from 116 fixed radar sensors in the Bay Area, demonstrating that such variance-based features often correlate strongly with flow. Similarly, \citet{neumann2013dynamic} proposed a dynamic Bayesian network that uses speed data collected from a fleet of 4{,}300 probe taxis in Berlin, revealing improved accuracy when excluding freeways and highlighting how temporal dependencies shape flow estimation.

In the context of urban roads, \citet{kumarage2018traffic} demonstrated that crowdsourced travel time data from the Google Distance Matrix API, combined with archived flow records and geometric features, can yield cost effective estimates of traffic flow. \citet{pun2019multiple} further underscored the importance of integrating topological measures (e.g., Degree, Betweenness, Closeness) alongside geometrical properties like road length, showing that multiple regression models, particularly Random Forest and multiple linear regression can outperform simpler single-measure approaches for most datasets. Meanwhile, \citet{li2021multi} introduced a \gls{gpr} to reconstruct traffic flows from \gls{fcd}, concluding that a multi-model strategy (where regressors are tailored to different types of days) surpasses a single-model approach.

Beyond classical machine learning approaches, ensemble learning methods have garnered increasing attention. \citet{sun2023improved} applied LightGBM and XGBoost, each optimized via Bayesian hyperparameter tuning, and found that these integrated models significantly outperformed the Van Aerde model \citep{rakha2002comparison}, random forests, and backpropagation neural networks. Notably, XGBoost delivered particularly strong performance on expressways, while LightGBM performed better in main and secondary roads. Further illustrating the growing reliance on indirect speed-to-flow mapping, \citet{mahajan2023predicting} used exogenous variables (e.g., speed, road attributes) and transfer learning on an \gls{lstm} architecture, enabling models trained in data rich regions to be finetuned for data scarce locations.

Taken together, these contributions highlight the persistent complexity of estimating traffic flow from more readily obtained indicators such as speed. They also show how factors like road geometry, topological connectivity, and exogenous contextual data can help refine predictions, but cannot fully resolve the complexities that arise from the nonlinearities of traffic states.

Although the methods reviewed above have advanced traffic flow estimation through classical machine learning and statistical tools, they often struggle to capture the large scale spatiotemporal patterns and network heterogeneity inherent in real-world deployments. Moreover, many traditional approaches are limited when attempting to extrapolate traffic flow at locations without direct measurements. While traditional spatiotemporal imputation (kriging) methods and related geostatistical techniques do offer an inductive way to estimated unobserved locations' values, they may not fully exploit the graph like connectivity inherent in road networks.

Motivated by these challenges, recent advances in \gls{gnn} have opened new avenues for analyzing spatiotemporal data, as they explicitly model complex interactions both across time and between nodes (e.g., nearby traffic sensors) \citep{wu2021inductive}. According to a survey by \citet{jin2024survey}, \gls{gnn} based methods have demonstrated substantial improvements over traditional analytic tools such as \gls{svr}, \gls{var}, and \gls{arima} by explicitly modeling intricate spatial temporal dependencies in timeseries data. These models broadly fall into inductive and transductive learning paradigms. 

Transductive \gls{gnn}s like GACN \citep{ye2021spatial} require the full graph during training process. They directly learn node specific embeddings, which limit their applicability to static and predefined graphs. This family of models are useful for imputation of missing data on fixed locations due to sensor failures, and maintenance periods, however they cannot be generalized to new nodes.
In contrast, inductive GNNs learn generalized aggregation functions that enables them to generate embeddings dynamically for unseen nodes or entirely new graph structures without retraining \citep{hamilton2017inductive, wu2021inductive}.

In the context of spatiotemporal imputation (kriging), inductive \gls{gnn}s exhibit substantial practical advantages due to their capability to accommodate dynamic changes in sensor networks (e.g., sensor failures, relocation, or new deployments)  without retraining from scratch \citep{wu2021inductive,zheng2023increase,xu2023kits,wei2024inductive}.
In addition to dynamic changes in sensor networks, inductive models are beneficial to make estimations where sensor installation is not practical due to costs or other barriers.

However, while inductive approaches have demonstrated robust performance in spatial imputation tasks involving variables such as traffic speed, they have shown limited effectiveness in imputing traffic flow data. Traffic speed, being a relatively smooth and continuous phenomenon, aligns well with assumptions implicit in inductive modeling approaches. In contrast, traffic flow exhibits noticeable variability across nearby sensors, often differ significantly due to structural and functional distinctions (e.g, ramps vs. main road segments). This heterogeneity presents a substantial challenge for inductive methods, as reflected in empirical studies and our benchmarks. \citet{xu2023kits} propose a novel inductive spatio‐temporal kriging framework named \textbf{KITS} that leverages an Increment training strategy. They conducts benchmarks on 3 traffic speed datasets
and a traffic flow dataset
between their proposed methodology KITS, and Mean aggregation, OKriging \citep{cressie2011statistics}, KNN, KCN \citep{appleby2020kriging}, IGNNK \citep{wu2021inductive}, LSJSTN \citep{hu2023decoupling}, and INCREASE \citep{zheng2023increase}.
Their model outperform all the mentioned dataset and they report promising \gls{mape} values for traffic speed datasets, whereas substantially higher errors were observed for traffic flow imputation.

This discrepancy arises from a fundamental challenge in graph representation learning known as heterophily, where connected nodes in a graph, contrary to the common GNN assumption of homophily, exhibit dissimilar features or belong to different classes \citep{zheng2022graph}. In road networks, this is particularly pronounced, where a high-capacity motorway segment and its adjacent low-capacity exit ramp are directly connected but possess vastly different flow characteristics. Standard \gls{gnn}s, which often act as low-pass filters by averaging neighbor information, struggle in such settings as the aggregation of features from dissimilar neighbors can corrupt the node's representation, leading to poor predictive performance \cite{zhu2023heterophily}. Figure \ref{fig:flow_comparison_1620_14611} depicts traffic flow in two nearby loop detectors in Antwerp where the heterophily in the scale of traffic flow is visible, along with the homophily on the traffic flows' trends.

Given these limitations of purely inductive methods for flow, there is a clear need for methodologies that can impute traffic flow more accurately while still leveraging the generalization capabilities needed for unobserved locations. Table~\ref{tab:model_comparison} summarizes how existing state-of-the-art inductive kriging models address these challenges, and highlights the unique positioning of our proposed approach.

\begin{table}[H]
\centering
\caption{Comparison of model mechanisms, heterogeneity handling, inputs, and learning paradigm.}
\label{tab:model_comparison}
\resizebox{\textwidth}{!}{
\setlength{\tabcolsep}{6pt}
\renewcommand{\arraystretch}{1.2}
\begin{tabular}{L{3.1cm} L{3.2cm} L{5.4cm} L{4.0cm} L{3.3cm}}
\toprule
\textbf{Model} & \textbf{Core Mechanism} & \textbf{Heterogeneity Strategy} & \textbf{Input Data} & \textbf{Learning Paradigm} \\
\midrule
IGNNK \citep{wu2021inductive} & GCN-based & Implicit (via attention) & Flow only & Fully inductive \\
KITS \citep{xu2023kits} & Spatiotemporal Conv. & Incremental training & Flow only & Fully inductive \\
IAGCN \citep{wei2024inductive} & Adaptive GCN & Adaptive graph structure & Flow only & Fully inductive \\
\addlinespace[2pt]
\textbf{Ours} & \textbf{Transformer + GCN-FiLM} & \textbf{Explicit (FiLM, Node Scaling)} & \textbf{Flow, Speed, OSM, Simulation} & \textbf{INDU-TRANSDUCTIVE} \\
\bottomrule
\end{tabular}
}
\end{table}

Motivated by this, this paper introduces an \textbf{INDU-TRANSDUCTIVE} training strategy. This approach is designed to combine paradigms: it leverages the relative abundance and smoother nature of speed data using a transductive perspective with the assumption the traffic speed can be obtained or estimated network wide using \gls{fcd} or simpler models, while applying inductive learning specifically to the more challenging, heterogeneous and harder to measure flow variable. This allows the model to generalize flow predictions to new locations (inductive strength) while being anchored and informed by the more stable, network wide speed patterns and external features. In doing so, our framework aims to fuse the strengths of both paradigms enabling inductive imputation for flow while simultaneously learning the complex relationship between external node features, observed speed patterns, and the target flow variables.

Specifically, our methodology adopts a transductive modeling approach for traffic speed (where we assume it is cheaper to acquire from other source such as navigation applications) and an inductive approach for traffic flow (that is challenging to acquire using navigation applications) to take advantage of on inductive capabilities to impute flow variable. Furthermore, we integrate best practices from recent inductive models in our model. Detailed model architecture and the training strategy are outlined in Section ~\ref{sec:proposed_framework}.

This paper contributes the following:
\begin{itemize}
    \item Introduces the INDU-TRANSDUCTIVE strategy, a novel hybrid GNN training approach that is tailored for the specific challenges of traffic flow imputation.

    \item Leverages static calibrated and non calibrated traffic simulation to improve the Kriging task's accuracy.

    \item Leverages readily available or easily estimated speed data transductively across the network, while applying inductive learning specifically to the sparse, hard to get, and heterogeneous flow variable.
    \item Integrates external geospatial features derived from \gls{osm} and employs graph structures from driving distances to effectively capture network heterogeneity and functional differences between road segments.
    \item Demonstrates through extensive experiments on multiple real world datasets that the proposed approach significantly outperform state-of-the-art inductive baselines for traffic flow imputation at unobserved locations.

\end{itemize}

\section{Dataset}
\label{sec:dataset}
\subsection{MOW}
The main dataset used in this study, referred to as \gls{mow}, was provided by the Roads and Traffic Agency (Agentschap Wegen en Verkeer). It consists of traffic measurements collected via loop detectors deployed across Antwerp, Belgium.

The dataset spans the period from January 1, 2022, to January 1, 2023, and include measurements from  207 unique locations. Among these, 105 locations are classified as \textbf{motorway} and 102 are categorized as \textbf{motorway\_link} based on the \gls{osm} network classification.

Traffic flow is represented using \gls{pce}. The average speed is calculated by taking the mean of speed measurements across all lanes at each location and in 15-minute intervals. Similarly, traffic flow is calculated as the sum of \gls{pce} values across all lanes, aggregated into 15-minutes intervals.

The dataset exhibits an average speed of approximately 81 km/h across all locations, while the average traffic flow is around 468 \gls{pce} per 15 minutes, equivalent to 1872 \gls{pce} per hour. These statistics provide an overview of typical traffic conditions across the monitored locations.

The time series data is partitioned chronologically into three segments:
\begin{itemize}
    \item Training set (60\%): used to train the proposed model and baseline models.
 \item Validation set (20\%): used to tune hyperparameters and prevent overfitting.
 \item Test set (20\%): used for final model evaluation.
\end{itemize}
   
\begin{figure}
\centering
\includegraphics[width=1.\textwidth]{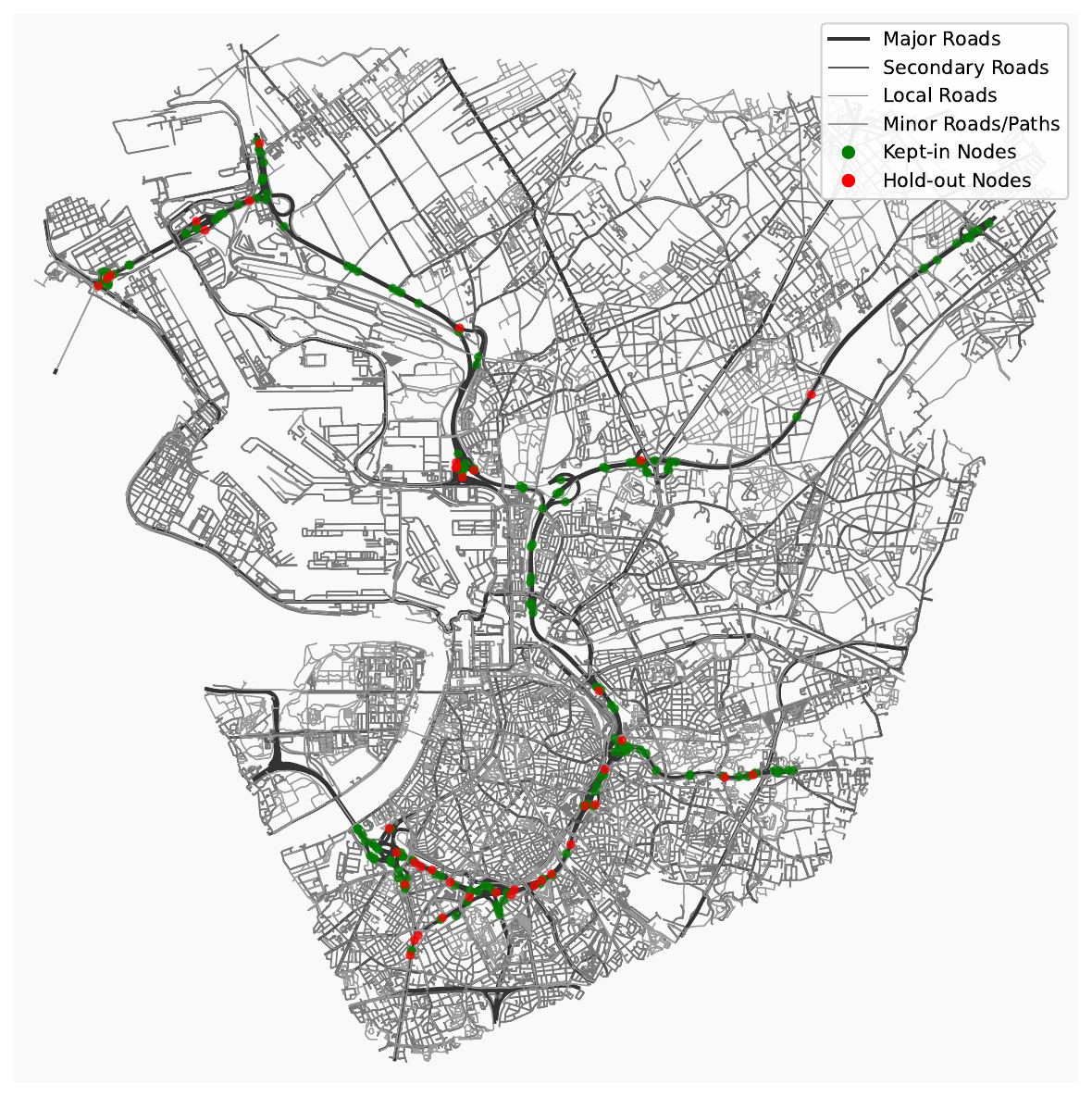}
\caption{\gls{mow} loop detector locations used in this study.}
\label{fig:MOW_locations}
\end{figure}

Figure \ref{fig:MOW_locations} shows the locations of the loop detectors for the MOW dataset used in this study. Green dots represent the nodes where the model is trained using observed traffic flow, while red dots indicate nodes where traffic flow values are masked during training. We later refer to these as hold-out nodes.

We used Open Source Routing Machine \cite{luxen-vetter-2011} to calculate the driving distance between each pair of locations using Multi-Level Dijkstra's algorithm, then the distances were normalized and top 5 connections were kept as our adjacency matrix. For the benchmark models, we experimented with varying numbers of kept connections to ensure a fair comparison.

\begin{figure}
\centering
\includegraphics[width=1.\textwidth]{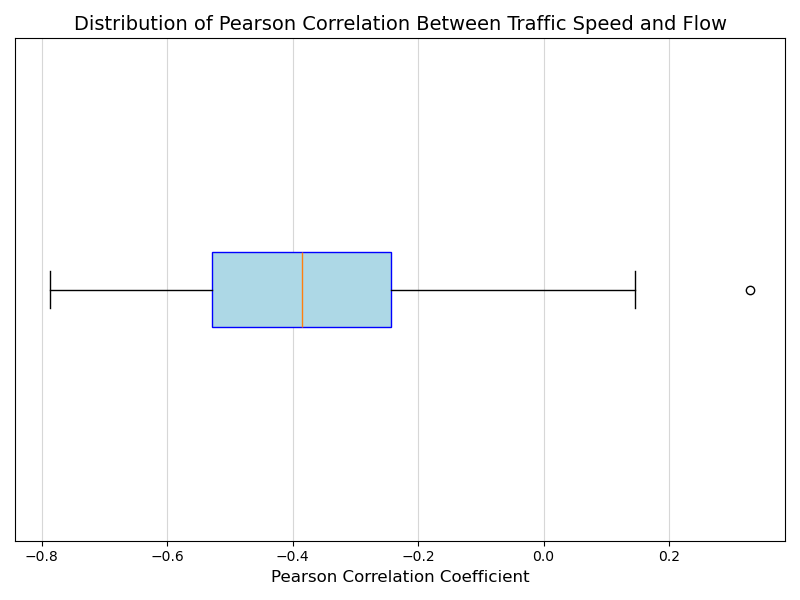}
\caption{Pearson correlation values boxplot}
\label{fig:Pear_locs_boxplot}
\end{figure}

\begin{sidewaysfigure}
\centering
\includegraphics[width=\textwidth]{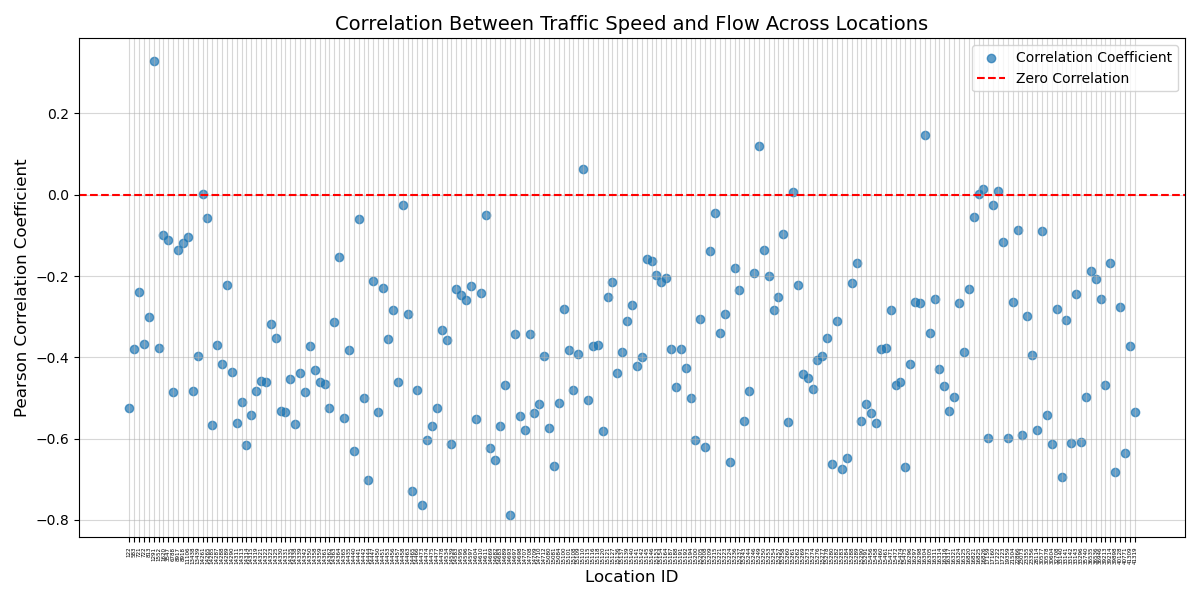}
\caption{Pearson correlation value between traffic speed and flow for all locations}
\label{fig:Pear_locs}
\end{sidewaysfigure}

To better understand the relationship between traffic speed and flow, a Pearson correlation analysis was conducted across all locations. The results are summarized in Figures \ref{fig:Pear_locs_boxplot} and \ref{fig:Pear_locs}.
\begin{itemize}
    \item Median Correlation Coefficient: Approximately -0.4, indicating a moderately negative correlation between traffic speed and flow.
\item \gls{iqr}: Ranges from -0.6 to -0.2, showing that most correlations fall within this range.
\end{itemize}
This negative correlation aligns with expectations, as increased traffic flow generally leads to a decreased traffic speed. However, the observed correlations are moderate to weak, suggesting that the relationship between traffic speed and flow varies significantly across different locations.

\begin{figure}
    \centering
    \begin{subfigure}{0.45\textwidth}
        \centering
        \includegraphics[width=\textwidth]{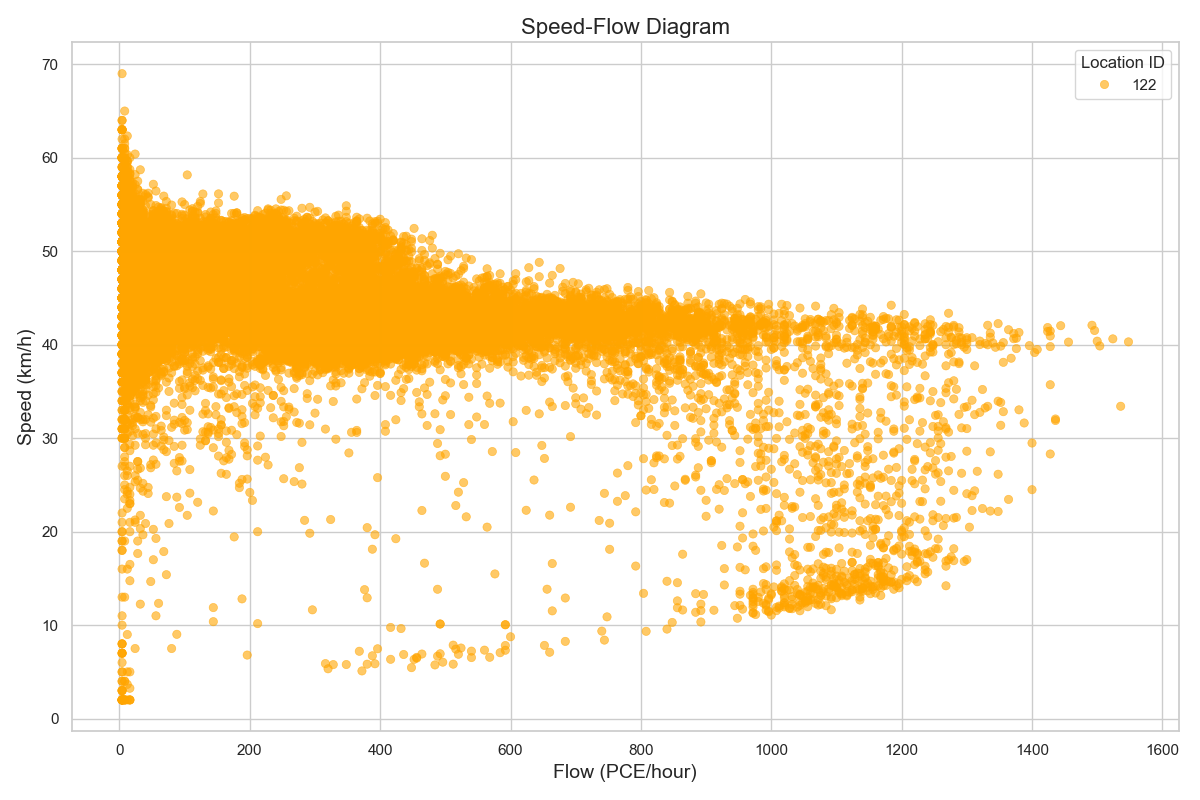}
        \caption{Motorway Link (Location ID 122)}
        \label{fig:speed_flow_122}
    \end{subfigure}
    \hfill
    \begin{subfigure}{0.45\textwidth}
        \centering
        \includegraphics[width=\textwidth]{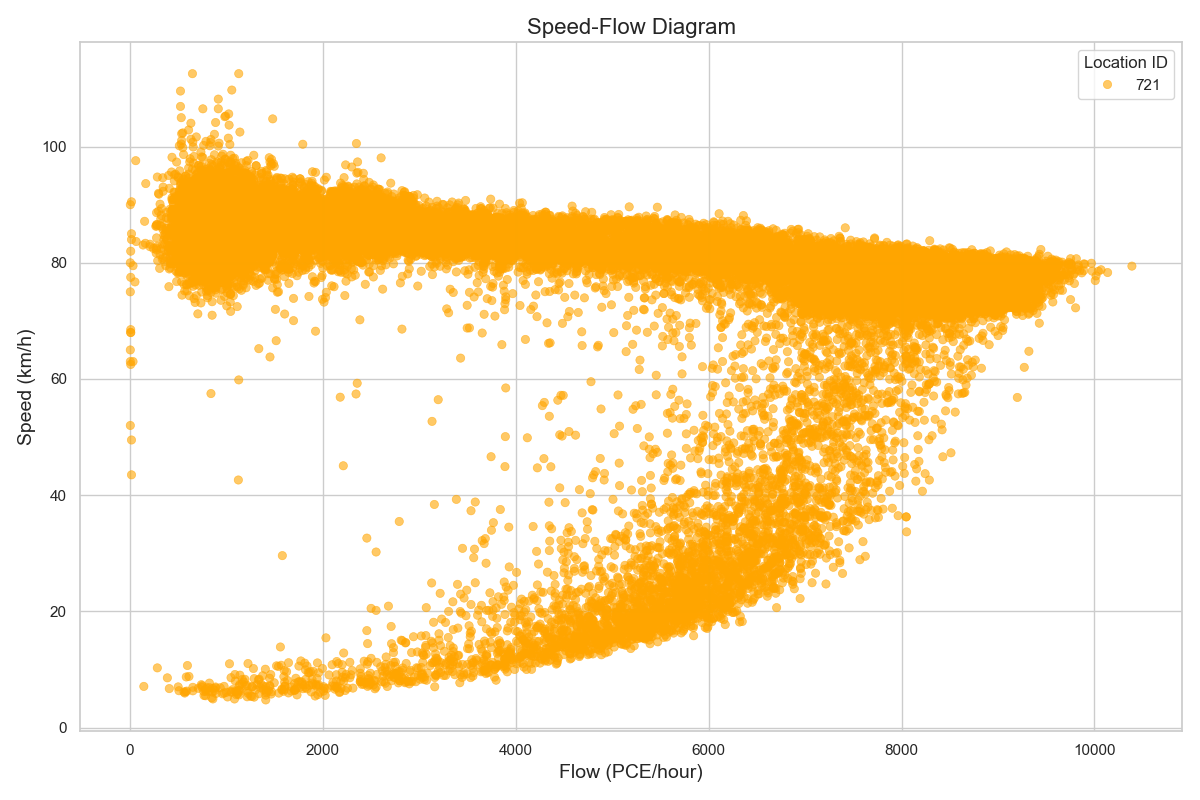}
        \caption{Motorway (Location ID 721)}
        \label{fig:speed_flow_721}
    \end{subfigure}
    \caption{Speed-Flow Diagrams: Comparison between a motorway (Location ID 721) and a motorway link (Location ID 122). They demonstrate high speed stability even at higher traffic flows. These diagrams are plotted based on the whole duration of the dataset.}
    \label{fig:speed_flow_diag}
\end{figure}

Figure \ref{fig:speed_flow_diag} presents speed-flow diagrams for two selected locations from the loop detector dataset, illustrating different traffic dynamics. The motorway segment (Location ID 721) demonstrates relatively stable traffic speeds, even at higher traffic flows, suggesting that factors like speed limits, rather than traffic density, play a dominant role in constraining free-flow speeds. On the other hand, the motorway link (Location ID 122) shows a more pronounced decline in speed as traffic flow increases, reflecting a capacity driven relationship.

These findings highlight the challenges of using speed as a direct proxy for traffic flow. While the expectation is that higher traffic flows generally reduce speeds, the actual relationship is weakened by the fact that free flow speeds are not entirely determined by volume. Instead, legal limits and other external factors limit the maximum achievable speed. As a result, traffic flow estimation based solely on speed data becomes challenging. In many cases, increased flow may not substantially reduce speed until the network approaches its physical thresholds. This diminishes the correlation strength and complicating the direct inference of flow from speed measurements. 

\begin{figure}
\centering
\includegraphics[width=1.\textwidth]{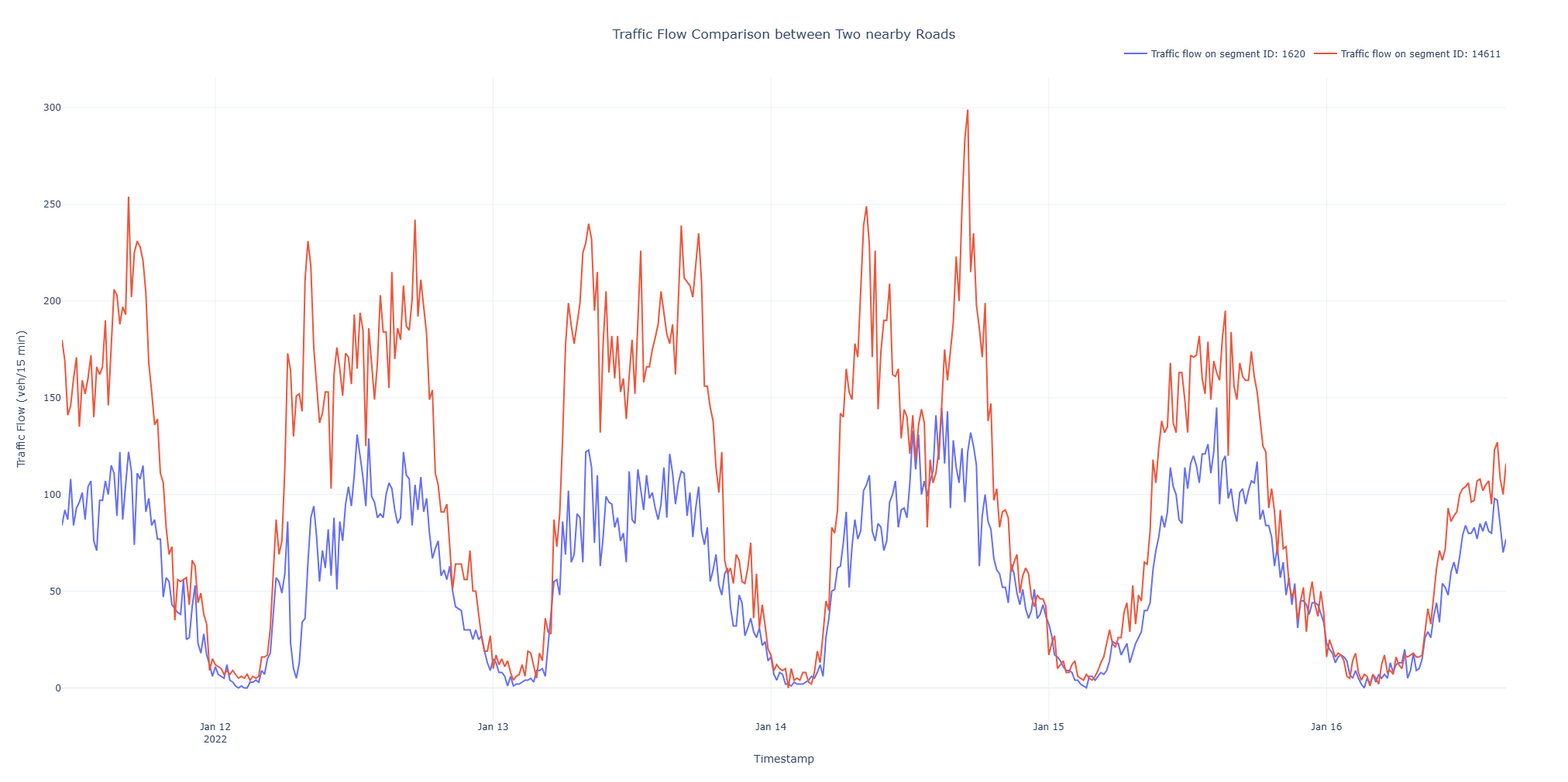}
\caption{Traffic flow plot on two nearby sections. Even though the traffic flow mostly follows a similar trend, the scale varies significantly between the two.}
\label{fig:flow_comparison_1620_14611}
\end{figure}

Message passing in \gls{gnn}s typically assumes feature homophily, meaning that connected or nearby nodes often share similar characteristics.
We have observed that this assumption does not hold in road traffic flows.
To clearly demonstrate the substantial differences in traffic flow that can occur between nearby locations, Figure~\ref{fig:flow_comparison_1620_14611} compares traffic patterns on two adjacent road sections. As seen in the figure, traffic flows' scale differs significantly between nearby segments, even though they mostly follow the same trend.


\subsection{UTD19-Torino}

\begin{figure}
\centering
\includegraphics[width=1.\textwidth]{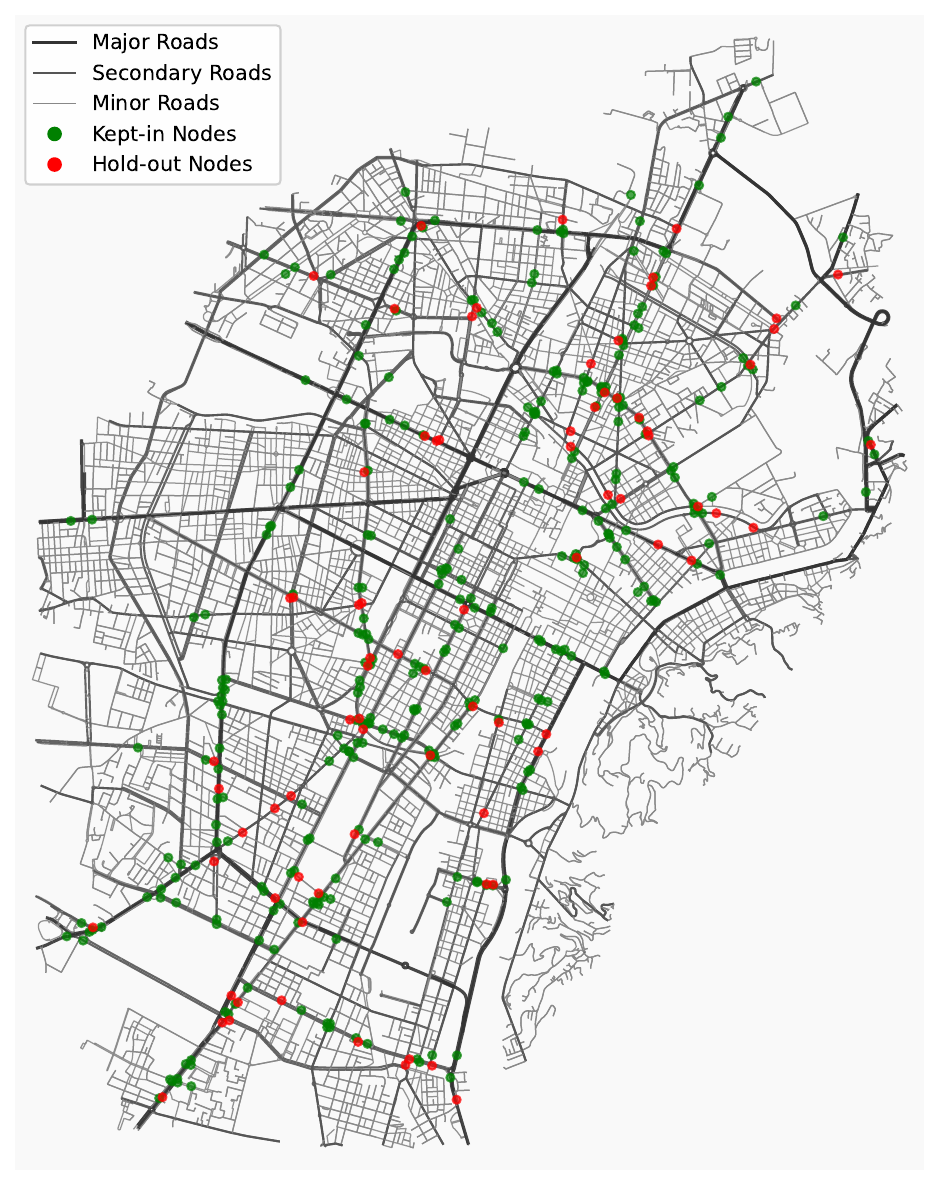}
\caption{UTD19-Torino loop detector locations used in this study.}
\label{fig:utd19Torino_locations}
\end{figure}

This dataset covers the period from 2016-09-26 00:00:00 to 2016-10-16 23:55:00 in 5-minute intervals. It has 399 locations, of which 320 are retained for model training while 79 are held out for testing. The average speed for this dataset is 42 km/h, while the average traffic flow is 408 Veh/h. Adjacency matrices are computed similarly to MOW, using driving distances. Similar to the MOW dataset, the time series data is split chronologically into 3 segments, where the first 60\% is kept for training, the next 20\% is kept for validation and the last 20\% for the testing. Figure \ref{fig:utd19Torino_locations} shows the locations of the loop detectors for the UTD-19 Torino dataset used in this study.

\subsection{UTD19-Essen}

\begin{figure}
\centering
\includegraphics[width=1.\textwidth]{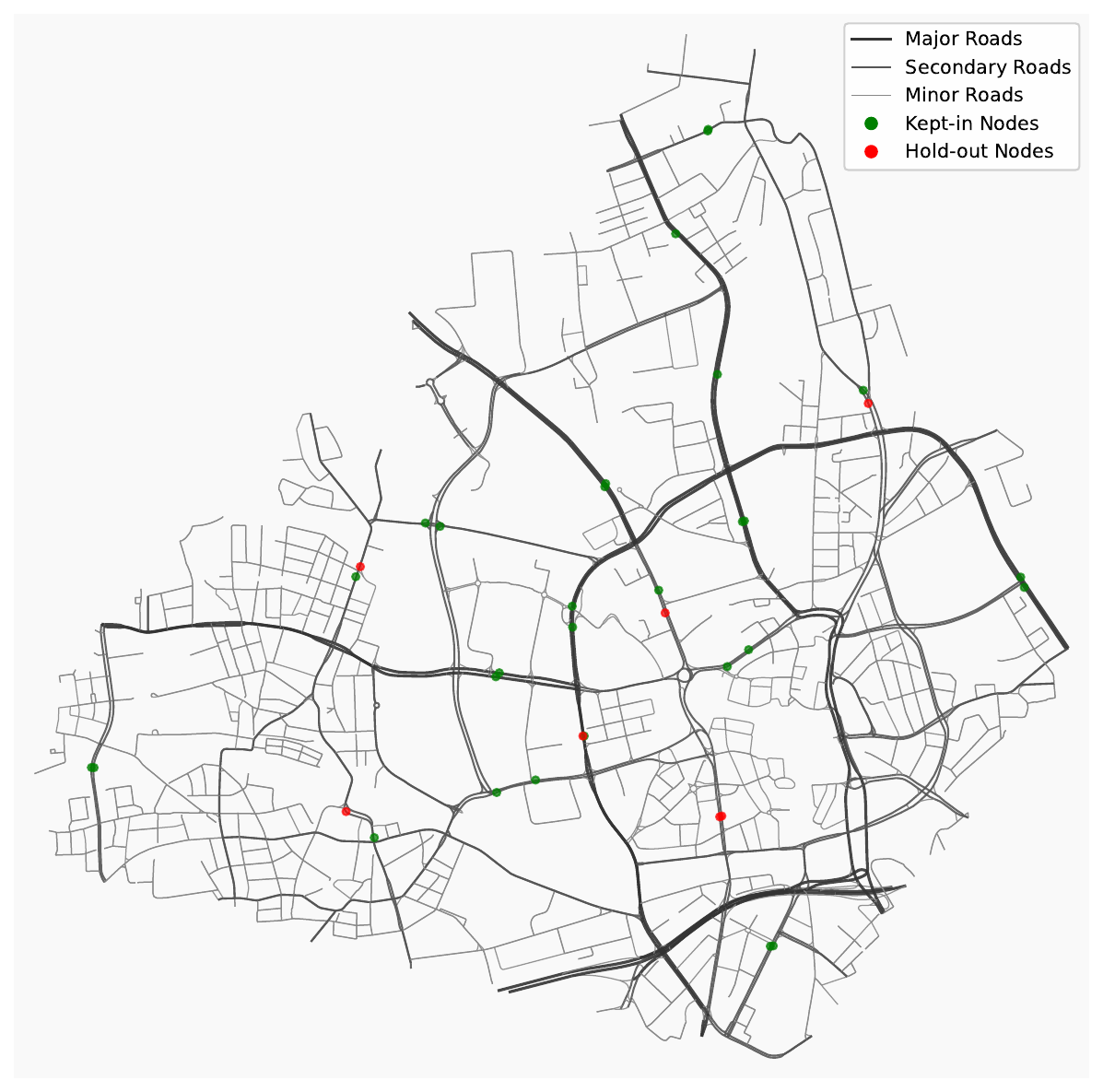}
\caption{UTD19-Essen loop detector locations used in this study.}
\label{fig:utd19Essen_locations}
\end{figure}

Spanning from 2017-09-01 00:25:00 to 2017-09-30 23:50:00 in 5-minute intervals, this dataset includes 36 locations, of which 29 are used for training and 7 are held out. The average speed for this dataset is 53 km/h, while the average traffic flow is 231 Veh/h. The adjacency matrix uses the same driving-distance-based approach. Similar to the MOW and UTD19-Torino datasets, the time series data is split chronologically into 3 segments, where the first 60\% is kept for training, the next 20\% is kept for validation and the last 20\% for the testing. Figure \ref{fig:utd19Essen_locations} shows the locations of the loop detectors for the UTD19-Essen dataset used in this study.

\section{Background}

\subsection{Acquisition and Processing of External Features}

To accurately capture the intricate spatiotemporal patterns influencing traffic flow, this study integrates external features extracted from \gls{osm}. We utilize OSMnx to systematically collect and preprocess node-specific features relevant for traffic analysis.

\subsubsection{Feature Acquisition}

For each sensor location defined by its geographical coordinates (latitude and longitude), we construct a local road network subgraph within a 500-meter radius. This process leverages OSM's metadata which enables extraction of detailed infrastructure and contextual features in a network-wide manner.

Specifically, the following features are collected for each sensor:

\begin{itemize}
    \item \textbf{Edge-Level Attributes}: attributes directly extracted from the nearest road segment such as \textit{lanes, highway type, maxspeed, length, and curvature}. Curvature is calculated as ratio of edge length to straight line distance between first and last coordinates and is an indicator for ramps.
    \item \textbf{Intersection Density}: computed as the count of intersections (nodes with degree $\geq 3$) normalized by the area of the convex hull enclosing the network around the sensor.
    \item \textbf{Average Betweenness Centrality}: calculated using NetworkX, weighted by road segment length, capturing the importance of nodes in terms of road network connectivity.
    \item \textbf{Traffic Signal Count}: number of nodes tagged as traffic signals within a 500 meter radius.
    \item \textbf{Predominant Land Use}: determined from OSM tags for land use categories within the sensor vicinity.
\end{itemize}

These attributes provide insights into local infrastructure characteristics and help the model distinguish different road types and their behaviors.

\subsubsection{Feature Processing}

Once acquired, external features undergo multiple preprocessing steps to transform them into numerical representations suitable for model training:

\begin{itemize}
    \item \textbf{Boolean and Categorical Encoding}: Boolean attributes (e.g., \textit{oneway}, \textit{bridge}, \textit{tunnel}) are converted into numerical binary values (1 for True, 0 for False). Categorical attributes like \textit{highway type} and \textit{land use} are transformed using one-hot encoding.
    \item \textbf{Numerical Normalization}: Continuous numerical variables (e.g., \textit{maxspeed, length, curvature}) are normalized between 0 and 1 to standardize feature scales and stabilize model training.
\end{itemize}

The feature extraction process includes robust error handling to account for incomplete or missing data within \gls{osm}. Finally, extracted features from each location are aggregated into a unified feature matrix. A summary of the collected and processed external features is provided in Table \ref{tab:feature_processing_summary}.

\begin{table}[H]
\centering
\caption{Summary of external features acquired from OSM.}
\label{tab:feature_processing_summary}
\resizebox{\textwidth}{!}{
\setlength{\tabcolsep}{8pt}
\renewcommand{\arraystretch}{1.25}
\begin{tabular}{L{3.6cm} L{8.2cm} L{4.2cm}}
\toprule
\textbf{Feature Category} & \textbf{Feature Name} & \textbf{Processing Technique} \\
\midrule
\multirow{4}{*}{Edge-level features}
& lanes, maxspeed, length & Normalization \\
& oneway, bridge, tunnel   & Binary encoding \\
& highway type             & One-hot encoding \\
& road curvature           & Normalization \\
\cmidrule(lr){1-3}
\multirow{2}{*}{Topological features}
& Intersection density          & Normalization \\
& Avg.\ betweenness centrality  & Normalization \\
\cmidrule(lr){1-3}
\multirow{2}{*}{Contextual features}
& Traffic signal count     & Count normalization \\
& Predominant land use     & One-hot encoding \\
\bottomrule
\end{tabular}
}
\end{table}

\subsubsection{Traffic flow simulation}

For cities of Antwerp, Essen and Turin, we began with an \gls{osm} extract for Essen and Turin in \gls{pbf} and convert it to appropriate format and then generate the road network with SUMO's network builder. Railways, non-motorized links, and isolated links were pruned to obtain a connected motor traffic network.

Synthetic travel demand was produced with SUMO's random trip generator for a 24 hour horizon. A time varying departure schedule was specified as a 24 element pattern that assigns a constant inter departure spacing to each successive were drawn from network edges with probabilities weighted by lane count and by edge speed. To better represent traffic dynamics, only trips exceeding 500m were allowed.

The microscopic simulation was run with a one-second step on these networks using the routed demand described above, for the 24 hour period. The results are aggregated to represent hourly traffic flow for all of the road's segments.

The simulation described above is not accurate in itself. The idea is that even with this inaccuracy, some information can be extracted from the relationship among the nodes. For example, the main arterial roads exhibit more traffic flow compared to a residential area. We hypothesize that the proposed deep learning model can extract relationship between nodes, even though the simulation is imperfect.

In addition to the basic simulation described above, the Antwerp case was supplemented with link-level hourly flows from a calibrated \gls{sta} described in \cite{mahdaviabbasabad2025hybrid}. That \gls{sta} was built on a processed \gls{osm} network and an hourly OD matrix from the Flemish Government. A stochastic user equilibrium was computed with \gls{bpr} link costs, detailed intersection delays such as signals, right of way, and roundabouts were added, and spillback was modeled via the node model proposed by \cite{tampere2011generic}. The \gls{sta} was calibrated against several data sources available in the area. The model trained using this simulation is referred to as "Proper simulation" in the following sections. Detailed descriptions of the model formulation, calibration, and validation are provided in \cite{mahdaviabbasabad2025hybrid,corthout_opbouw_2022}.

\section{Proposed Framework}
\label{sec:proposed_framework}

We consider a road network as a graph $G = (V, E)$, represented by an adjacency matrix $A \in \mathbb{R}^{N \times N}$, where $V$ is a set of nodes (traffic sensors) and $E$ is the set of edges representing connectivity between nodes. Each node $v \in V$ records time series of traffic speed and flow measurements, denoted by $X^{(\mathrm{speed})}$ and $X^{(\mathrm{flow})}$ respectively. The data from all sensors over a time window is organized as a matrix $X \in \mathbb{R}^{N \times T}$ (with $N=|V|$ sensors and $T$ time steps). Inputs are normalized via an operator $\mathcal{N}$. Additionally, static node attributes $X_{\text{static}}$ and temporal features (e.g., time-of-day) $X_{\text{time}}$ are provided. Traffic flow imputation is the task of estimating the unobserved or missing entries in the flow matrix using the observed entries and the graph structure.
We denote the training, validation, and test datasets as $\mathcal{D}_{\text{train}}$, $\mathcal{D}_{\text{val}}$, and $\mathcal{D}_{\text{test}}$, respectively, consisting of batches of observed sequences.

Formally, we designate a subset of nodes \(\mathcal{H} \subset V\) as \textit{hold-out} sensors ($\sim$20\% of nodes) which are never seen during training. The training task is formulated as a supervised learning problem: the model is trained on the temporal sequences in \(\mathcal{D}_{\text{train}}\) excluding \(\mathcal{H}\), and must generalize to predict flow values on \(\mathcal{H}\) in \(\mathcal{D}_{\text{val}}\) and \(\mathcal{D}_{\text{test}}\).
Performance is evaluated by metrics such as \gls{mae}, \gls{mape}, \gls{smape} between the imputed and true values.

\subsection{INDU-TRANSDUCTIVE training strategy}

Motivated by the distinct challenges posed by traffic speed and flow data, we introduce a hybrid training strategy termed \textbf{INDU-TRANSDUCTIVE}. Our approach simultaneously leverages transductive learning for the easier-to-obtain traffic speed and inductive learning for more heterogeneous and difficult to acquire traffic flow.

Specifically, each input sequence is represented as a dual-channel tensor $\mathbf{X} \in \mathbb{R}^{B \times N \times T \times 2}$ (where $B$ is the batch size) that contains both speed and flow information with fixed channel order \([0]=\) speed and \([1]=\) flow. To emulate realistic deployment scenarios and avoid overfitting, at each training epoch we randomly select a subset of nodes (road segments) based on a sampling ratio. For the selected nodes, the flow measurements are retained, whereas the flow values of unsampled nodes and hold-out sensors are masked to simulate a missing-data scenario during training. Speed remains available at all nodes (transductive side).

\subsubsection{Hard-node mining and the reconstruction loss}

Given the road network \(G=(V,E)\) with \(|V|=N\) sensors, \(|E|\) edges, and a time window of length \(T\), let
\(X^{(\mathrm{flow})},\,X^{(\mathrm{speed})}\in\mathbb{R}^{N\times T}\) denote the normalized flow and speed, and let
\(\widehat{X}^{(\mathrm{flow})}\) be the model output for flows when the dual-channel input
\(\mathrm{stack}\!\bigl(X^{(\mathrm{speed})},\,X^{(\mathrm{flow})}\bigr)\) is provided.
We define the training node set \(V_{\text{train}} := V\setminus\mathcal{H}\) with \(N_{\text{train}} = |V_{\text{train}}|\).
In each epoch \(e\), we draw a visibility ratio \(r_e\sim\mathcal{U}(r_{\min},r_{\max})\) using a uniform distribution with predefined boundaries, and select a
visible node set \(\mathcal{S}_e\subseteq V_{\text{train}}\) with \(|\mathcal{S}_e|=\lfloor r_e N_{\text{train}}\rfloor\).
Its complement in \(V_{\text{train}}\), \(\mathcal{R}_e=V_{\text{train}}\setminus\mathcal{S}_e\), is used for reconstruction.

While the training objective employs \gls{mae} (as part of a combined loss; see below), for hard node mining we deliberately adopt \gls{smape} to avoid a selection bias toward high-flow nodes.
Let $\mathcal{T}_{\text{val}}$ denote the set of time indices available in the validation set. The per-node difficulty $d_v$ is calculated as:
\[
d_v \;=\; \frac{1}{|\mathcal{T}_{\text{val}}|}\sum_{t\in\mathcal{T}_{\text{val}}} 
\underbrace{\frac{2\bigl|\mathcal{N}^{-1}(\widehat{X}^{(\mathrm{flow})}_{v,t})-\mathcal{N}^{-1}(X^{(\mathrm{flow})}_{v,t})\bigr|}
{\bigl|\mathcal{N}^{-1}(\widehat{X}^{(\mathrm{flow})}_{v,t})\bigr|+\bigl|\mathcal{N}^{-1}(X^{(\mathrm{flow})}_{v,t})\bigr|+\varepsilon}}_{\text{sMAPE}(v,t)}\!,
\qquad v\in V\setminus\mathcal{H},
\]
with a small constant \(\varepsilon>0\) to avoid division by zero. Note that we compute difficulties on denormalized data to ensure scale invariance on physical quantities.

This choice is scale-aware and approximately invariant to multiplicative rescaling:
if \(\widehat{X}^{(\mathrm{flow})}_{v,t}=X^{(\mathrm{flow})}_{v,t}\,(1+\rho_{v,t})\) with small relative
error \(\rho_{v,t}\), then
\[
\mathrm{MAE}\;\approx\;|\rho_{v,t}|\,\bigl|X^{(\mathrm{flow})}_{v,t}\bigr|
\]
grows linearly with the mean flow and over-selects large-flow sensors, whereas
\[
\mathrm{sMAPE}(v,t)
\;=\;
\frac{2\,|\rho_{v,t}|}{\,\bigl|1+\rho_{v,t}\bigr|+1+\varepsilon/\bigl|X^{(\mathrm{flow})}_{v,t}\bigr|\,}
\;\approx\;
\frac{2\,|\rho_{v,t}|}{\,2+\rho_{v,t}+\varepsilon/\bigl|X^{(\mathrm{flow})}_{v,t}\bigr|\,}
\;\approx\;
|\rho_{v,t}|,
\]
for \(|\rho_{v,t}|\ll 1\) and \(\bigl|X^{(\mathrm{flow})}_{v,t}\bigr|\gg\varepsilon\), which fairly compares nodes across heterogeneous scales.

Under \textbf{hard node mining}, reconstruction nodes are sampled without replacement from a tempered softmax over \(\{d_v\}_{v\in V_{\text{train}}}\):
\[
p_v \;=\; 
\frac{\exp(d_v/\tau)}{\sum_{u\in V_{\text{train}}}\exp(d_u/\tau)}\,,\qquad v\in V_{\text{train}},
\]
and we sample the reconstruction set \(\mathcal{R}_e\subseteq V_{\text{train}}\) without replacement according to \(\{p_v\}\), of size \(N_{\text{train}}-|\mathcal{S}_e|\), with \(\mathcal{S}_e = V_{\text{train}} \setminus \mathcal{R}_e\),
where \(\tau>0\), defined as the hard mining temperature, governs exploration-exploitation (smaller \(\tau\) concentrates mass on the hardest nodes). Let \(M_{\mathcal{S}_e},M_{\mathcal{R}_e}\in\{0,1\}^{N\times T}\) be binary masks that zero out entries outside \(\mathcal{S}_e\) or \(\mathcal{R}_e\), and let \(M_{\mathrm{keep}}\) zero out \(\mathcal{H}\). The epoch loss blends supervision on visible nodes and reconstruction on unseen nodes with a warm-up:
\[
\mathcal{L}_e \;=\; (1-\alpha_e)\,\mathcal{L}\!\big(\widehat{X}^{(\mathrm{flow})}, X^{(\mathrm{flow})};\, M_{\mathrm{vis}}\big)
\;+\;
\alpha_e\,\mathcal{L}\!\big(\widehat{X}^{(\mathrm{flow})}, X^{(\mathrm{flow})};\, M_{\mathrm{rec}}\big),
\]
\[
M_{\mathrm{vis}} := M_{\mathrm{keep}}\odot M_{\mathcal{S}_e}, 
\qquad
M_{\mathrm{rec}} := M_{\mathrm{keep}}\odot M_{\mathcal{R}_e},
\qquad
\alpha_e=\min\!\left(\frac{e}{E_{\text{warm}}},1\right).
\]
where \(\mathcal{L}\) includes masked \(\mathrm{MAE}\) and $E_{\text{warm}}$ is the number of warm-up epochs during which the reconstruction loss weight $\alpha_e$ linearly increases.

\subsubsection{Noise injection during the training}

To model heteroscedastic sampling variability in \(X^{(\mathrm{flow})}\) and to prevent a degenerate identity mapping on visible nodes under the combined objective above, we inject \emph{Poisson} corruption on the \emph{visible-node input channel} prior to forming the dual-channel tensor, while evaluating the loss against the clean target. Let \(\mathcal{N}\) and \(\mathcal{N}^{-1}\) denote the normalization/denormalization operators. For each epoch \(e\), define a linear decay
\[
\delta_e \;=\; \max\!\left(0,\,1-\frac{e}{E_{\text{noise}}}\right),
\]
where $E_{\text{noise}}$ represents the number of epochs over which the noise injection is annealed to zero, and let \(\sigma>0\) be the noise scale. During each training step, for \(v\in\mathcal{S}_e\) and \(t\in\{1,\dots,T\}\), we write the denormalized flow as
\(\lambda_{v,t}=\mathcal{N}^{-1}\!\big(X^{(\mathrm{flow})}_{v,t}\big)\), and sample Poisson counts
\[
\tilde{Z}_{v,t}\sim\mathrm{Poisson}\!\big(\lambda^{+}_{v,t}\big),\qquad
\eta_{v,t}=\tilde{Z}_{v,t}-\lambda_{v,t},\qquad
\lambda^{\text{(in)}}_{v,t}=\big[\lambda_{v,t}+\sigma\,\delta_e\,\eta_{v,t}\big]_+,
\]
with \(\lambda^{+}_{v,t}=\max(\lambda_{v,t},0)\) and \([\cdot]_+=\max(\cdot,0)\). The noisy visible-channel input becomes
\[
\widetilde{X}^{(\mathrm{flow,in})}_{v,t} \;=\; \mathcal{N}\!\big(\lambda^{\text{(in)}}_{v,t}\big),
\]
whereas the target in the loss remains the clean \(X^{(\mathrm{flow})}_{v,t}\). This breaks the trivial solution in which the model simply copies the flow input on \(\mathcal{S}_e\): if the model were to implement the identity map on visible nodes, its expected error would satisfy
\[
\mathbb{E}\!\left[\big|\widetilde{X}^{(\mathrm{flow,in})}_{v,t}-X^{(\mathrm{flow})}_{v,t}\big|\right]
\;=\;\mathbb{E}\!\left[\,\big|\mathcal{N}\!\big(\sigma\,\delta_e\,\eta_{v,t}\big)\big|\,\right]\;>\;0,
\]
and, under the normal approximation to the Poisson, 
\(\mathbb{E}[|\eta_{v,t}|]\approx \sqrt{\tfrac{2}{\pi}}\sqrt{\lambda_{v,t}}\), yielding a strictly positive penalty that encourages denoising via spatiotemporal structure (i.e., kriging-like reconstruction) rather than mere copying. Because the corruption is restricted to \(\mathcal{S}_e\) and \(\delta_e\to 0\) as \(e\) increases, the augmentation injects realistic early-epoch variability for robustness while preserving late-epoch stability and the integrity of masked (\(\Omega^c\)) and hold-out (\(\mathcal{H}\)) supervision.

Optimization of the proposed model parameters is performed using Adam optimizer in conjunction with gradient clipping and learning rate scheduling to maintain numerical stability and stable convergence. The training process is summarized in Algorithm~\ref{alg:indu_transductive_training}.

\begin{algorithm}[t]
\caption{INDU\textendash TRANSDUCTIVE Training Strategy}
\label{alg:indu_transductive_training}
\begin{algorithmic}[1]
\Require Training loader \(\mathcal{D}_{\text{train}}\); validation loader \(\mathcal{D}_{\text{val}}\); test loader \(\mathcal{D}_{\text{test}}\);
adjacency \(A\); static node features \(X_{\text{static}}\); time features \(X_{\text{time}}\); model \(M\); optimizer \(O\); scheduler \(S\) (cosine);
max epochs \(E_{\max}\); early stopping patience \(P_{\max}\);
node sampling bounds \([r_{\min}, r_{\max}]\);
hold-out node set \(\mathcal{H}\); mining temperature \(\tau\);
masked-loss warmup \(E_{\text{warm}}\); noise scale \(\sigma\); noise decay \(E_{\text{noise}}\).
\State \textbf{Init:} best validation loss \(\mathcal{L}_{\text{best}} \gets \infty\);\; patience \(p\gets 0\);\; difficulties \(d_v \gets 0\) for \(v\in V\setminus\mathcal{H}\)
\For{\(e = 1, \dots, E_{\max}\)} \label{line:epoch-start}
  \State Set \(M\) to \textbf{train} mode
  \State Sample visibility ratio \(r_e \sim \mathcal{U}(r_{\min}, r_{\max})\)
  \State Calculate target size \(K_{\mathrm{vis}} = \lfloor r_e N_{\text{train}} \rfloor\)
    \State Form probabilities \(p_v \propto \exp(d_v/\tau)\) over \(V\setminus\mathcal{H}\)
    \State Sample reconstruction set \(\mathcal{R}_e\) \emph{without replacement} of size \((|V|-|\mathcal{H}| - K_{\mathrm{vis}})\) from \(p_v\)
    \State Set \(\mathcal{S}_e \gets (V\setminus\mathcal{H})\setminus \mathcal{R}_e\)

  \State Construct masks \(M_{\mathcal{S}_e}, M_{\mathcal{R}_e}, M_{\mathrm{keep}} \in \{0,1\}^{N \times T}\) by broadcasting indicators for \(\mathcal{S}_e, \mathcal{R}_e, V\setminus\mathcal{H}\)
  \State \(\alpha_e \gets \min(e/E_{\text{warm}},\,1)\), \quad \(\delta_e \gets \max(0,\,1-e/E_{\text{noise}})\)
  \For{each batch \((X^{(\mathrm{speed})}, X^{(\mathrm{flow})}, X_{\text{time}})\) in \(\mathcal{D}_{\text{train}}\)}
    \State Set target \(Y \gets X^{(\mathrm{flow})}\)
    \State \textbf{// Build dual-channel input and apply masks}
    \State \(\mathbf{X} \gets \mathrm{stack}(X^{(\mathrm{speed})}, X^{(\mathrm{flow})})\) \Comment{Shape: \(B \times N \times T \times 2\)}
    \State Zero the \emph{flow} channel on \(\mathcal{R}_e \cup \mathcal{H}\): \(\mathbf{X}[\cdot, \mathcal{R}_e \cup \mathcal{H}, \cdot, 1] \gets 0\)
    \State Calculate denormalized flow: \(\lambda \gets \mathcal{N}^{-1}(X^{(\mathrm{flow})})\)

      \State Sample Poisson: \(\tilde{Z}\sim \mathrm{Poisson}(\lambda^{+})\); set \(\eta \gets \tilde{Z}-\lambda\)
      \State \(\lambda^{\text{(in)}} \gets [\lambda + \sigma\,\delta_e\,\eta]_+\); \(\widetilde{X}^{(\mathrm{flow,in})} \gets \mathcal{N}(\lambda^{\text{(in)}})\)
      \State Replace input flow on \(\mathcal{S}_e\): \(\mathbf{X}[\cdot, \mathcal{S}_e, \cdot, 1] \gets \widetilde{X}^{(\mathrm{flow,in})}[\cdot, \mathcal{S}_e, \cdot]\)

    \State \textbf{// Forward pass + losses}
    \State \(\widehat{Y} \gets M(\mathbf{X}, X_{\text{time}}, X_{\text{static}}, A)\)
    \State \(\mathcal{L}_{\mathrm{vis}} \gets \mathcal{L}(\widehat{Y}, Y;\, M_{\mathrm{keep}}\odot M_{\mathcal{S}_e})\)
    \State \(\mathcal{L}_{\mathrm{rec}} \gets \mathcal{L}(\widehat{Y}, Y;\, M_{\mathrm{keep}}\odot M_{\mathcal{R}_e})\)
    \State \(\mathcal{L}^{\text{(batch)}} \gets (1-\alpha_e)\,\mathcal{L}_{\mathrm{vis}} + \alpha_e\,\mathcal{L}_{\mathrm{rec}}\) \ \
    \State Backpropagate \(\nabla \mathcal{L}^{\text{(batch)}}\); clip gradients; optimizer step
    \State Step scheduler \(S\) \textbf{per batch}
  \EndFor
  \State Set \(M\) to \textbf{eval} mode
  \State \textbf{// Validation on hold-out \(\mathcal{H}\) and difficulty update}
  \State Evaluate \(\mathcal{L}_{\text{val}}\) and metrics on \(\mathcal{D}_{\text{val}}\)
  \State Compute node difficulties \(d_v\)
  
\EndFor
\State \textbf{// Testing}
\State Load best checkpoint; set \(M\) to \textbf{eval}; for each batch in \(\mathcal{D}_{\text{test}}\): mask flow on \(\mathcal{H}\) in input, forward, report metrics on \(\mathcal{H}\);
\end{algorithmic}
\end{algorithm}

\subsection{Proposed architecture}

We design a novel deep learning architecture that fuses inductive and transductive learning components for spatiotemporal Kriging in order to implement INDU-TRANSDUCTIVE strategy. In essence, the model treats traffic speed as a fully observed (transductive) feature across the entire network, while traffic flow is the target variable to be imputed inductively at unobserved locations. This is achieved by formulating a dual-channel input where each road segment has two features at each time step: its speed which is always provided and its flow where its only provided on the unmasked locations. By leveraging the abundant traffic speed information, and the simulation results, the model can infer flow more reliably. Below we outline the key components of the proposed \textbf{H}ybrid \textbf{IN}ductive-\textbf{T}ransductive Network, \textbf{HINT}. 

\begin{itemize}
    \item \textbf{Static Feature Encoder:} Each node (sensor locations) is associated with a vector of static features (e.g. road type, lane count, simulated flow in a normal weekday). We employ an External Feature Module consisting of a gated feed-forward network to embed these attributes into a latent vector per node. A learnable feature gate applies element-wise coefficients to selectively attenuate or amplify each input feature via a sigmoid layer. This gating is trained with an L1 regularization to encourage sparsity which ensures that only the most informative features contribute. The gated features then pass through two fully connected layers with SELU activations and batch normalization, yielding the static node embedding. This module helps the model capture heterogeneity caused by functional class differences (e.g., between highway segments and ramps) by using external features from \gls{osm} and simulation results.
    
    \item \textbf{Temporal Context Encoder:} We implemented a Time Features Module that encodes each time step's contextual variables into a latent time embedding to capture periodic temporal patterns such as hour of day, day of week, etc. This module is two-layer feed-forward network with ReLU activations and dropout applied to periodic cyclic encoding time features. It also includes a final gating layer that multiplies the output, enabling the network to recalibrate the importance of temporal features dynamically. The resulting time embeddings allow the model to account for daily and weekly cycles that affect traffic. These embeddings are broadcast across all nodes for a given time step to ensure every node's input with the same temporal context vector.
    
    \item \textbf{Inductive Spatial Transformer:} For each time step, the model performs an inductive spatial aggregation by treating the node embeddings as a sequence and applying a multi-head self-attention Transformer Encoder. Specifically, at time $t$, we construct a sequence of length $N$ (number of nodes), where each element is the sum of (i) the node's current dynamic features (obtained by projecting raw speed/flow inputs and time embedding through a linear layer) and (ii) the node's static embedding projected to the same dimension. This additive fusion injects static characteristics into the dynamic state of each node. The transformer's self-attention then operates on this sequence, allowing direct information exchange among all nodes without relying on a predefined adjacency matrix. This is crucial for inductive generalization: attention weights are learned as functions of node feature similarity, so the model can relate a new unseen node to other based on its features patterns (e.g. similar speed trends or static features). Unlike traditional GCN-based kriging which aggregates only fixed local neighbors, the inductive attention mechanism can capture long-range and context specific spatial dependencies across the network. In other words, this Transformer module plays a role similar to recent graph transformer approaches in traffic imputation and forecasting, providing a flexible learned notion of neighborhood that extends beyond static road distances.
    
    \item \textbf{Graph Convolution Branch:} In parallel with the transformer, our model includes a Graph Convolution pathway to exploit known physical connectivity. We feed each node's combined static-dynamic features (concatenation of static embedding and dynamic input features) into a multilayer Diffusion \gls{gcn} stack. This branch uses a precomputed adjacency matrix $A$ (based on shortest distance between nodes) to perform localized graph convolutions. Multiple \gls{gcn} layers are applied to propagate information across one-hop neighbors iteratively. To prevent the \gls{gcn} ignoring node-specific differences and explicitly address the heterophily challenge, we integrate a \gls{film} layer after each graph convolution \citep{perez2018film}: the \gls{film} takes the static node embedding as a conditioning input and produces a learned scale ($\gamma$) and shift ($\beta$) for each node's GCN output features, performing a feature-wise affine transformation. This \gls{film} conditioned \gls{gcn} allows the influence of neighbors to be modulated by each node's inherent properties. For example, a ramp's static features and simulation results can down-weight the traffic influence from a connected highway segment). This mechanism effectively introduces heterogeneity into the graph propagation, acknowledging that neighboring nodes may have functionally different traffic flow values. 
    
    \item \textbf{Multi-Source Fusion and Temporal Refinement:} The output of the inductive transformer branch and the \gls{gcn} branch represents complementary spatial perspectives where one is learned purely from data patterns and the other from the graph structure. We fuse these two representations using a learnable gating mechanism. Specifically, the model produces a pair of fusion weights through a small neural network that takes the concatenated transformer and \gls{gcn} feature vectors and outputs a softmax over two coefficients. This Dynamic Fusion layer adaptively weights the importance of the Transformer's signal versus the GCN's signals for each node and time. The weighted sum yields a unified spatial representation for each node at each time. Next, to capture temporal dynamics and ensure consistency over time, we pass the fused spatial features through a Temporal Refinement \gls{rnn}. This is a \gls{gru} that processes each node's sequence of fused features over the time window to smoothen short-term fluctuations and capture temporal dependencies. The \gls{gru} outputs a refined sequence which we then combine with the original transformer output and the fused output using a final Multi-Branch Fusion linear layer. This step allows the model to preserve both the initial spatial features and the refined temporal patterns. Finally, a readout fully-connected layer maps the fused feature vector of each node at each time to a single scalar prediction of traffic flow.
    
    \item \textbf{Node-wise Calibration/Scaling Layer:} A distinctive feature of our architecture is the NodeWise Scaling module that calibrates the model's output for each node. Given the final base prediction from the readout layer and the node's static embedding, this module predicts a multiplicative scale $\gamma_v$ and additive bias $\beta_v$ for each node $v$ and time step. The prediction is formulated as two simple linear layers applied to the static features (one each for $\gamma$ and $\beta$) and then $\gamma$ is passed through a softplus to ensure positivity (we initialize $\gamma$ and $\beta$ to be 1 and 0). The final flow estimate is $\hat{y}_{v,t} = \gamma_v \, y^{\text{base}} + \beta_v$. Intuitively, this acts like a learned per node affine calibration that allows the network to adjust for systematic biases. For example, if a residential area has consistently lower flow magnitude than what the shared layers predict from its neighbors and speed pattern, the scaler can downscale its outputs. This design addresses the spatial heterogeneity problem noted earlier, where even if nearby locations share trends, their absolute traffic flow levels can differ significantly. By learning these adjustments from data, the model ensures that the inductive predictions are anchored to each location's characteristics.
    
\end{itemize}

\begin{figure}
\centering
\includegraphics[width=1.\textwidth]{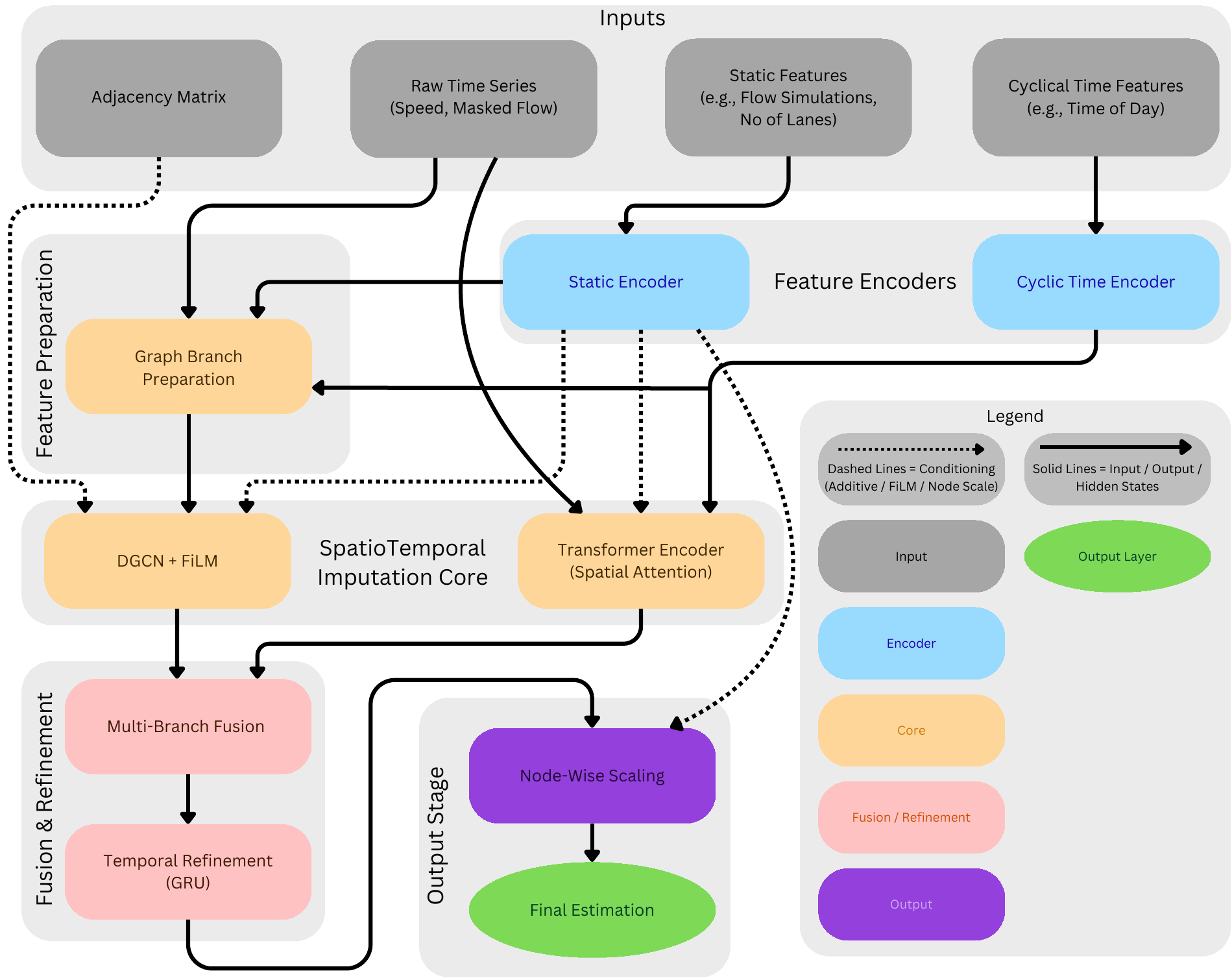}
\caption{Proposed architecture for INDU-TRANSDUCTIVE spatiotemporal traffic flow imputation. The model fuses transductive signals (traffic speed) and inductive targets (traffic flow) through static and temporal encoders, a Transformer-based inductive spatial attention module, and GCN-FiLM branch exploiting road network structure. Multi-branch Fusion with temporal refinement and node-wise calibration based on the external features and traffic flow simulation produces the final flow estimates.}
\label{fig:proposed_architecture}
\end{figure}

The outlined architecture is depicted in Figure~\ref{fig:proposed_architecture}.

\section{Results}

In this section we evaluate HINT against classical geostatistical methods (KNN, OKriging) and recent inductive GNN-based kriging models (IGNNK, IAGCN, KITS) on the three datasets introduced in Section~\ref{sec:dataset} (MOW, UTD19--Torino, UTD19--Essen). All models operate in the same inductive setting where during training and testing, flow values are observed at a sampled subset of sensors and must be imputed at unsampled sensors using the graph structure and any available covariates. Benchmark models only observe flow at sampled locations, whereas HINT additionally leverages network-wide speed, external OSM-based features, and, when available, simulated flows. We report \gls{mae}, \gls{rmse}, \gls{mape}, and \gls{smape} on the held-out sensors for each city (Table~\ref{tab:combined_metrics}).

The main quantitative results are summarized in Table~\ref{tab:combined_metrics}. Across all three datasets, HINT with basic simulation consistently outperforms KITS and other inductive baselines, with the largest relative gains on MOW and the smallest on UTD19--Essen. The row "Ablation-Without Simulation" isolates the contribution of simulation features: on MOW and UTD19--Torino, our model still outperforms KITS even without simulation, whereas on UTD19--Essen removing simulation degrades performance slightly below KITS.

For MOW, we additionally evaluate a calibrated "proper" simulation in addition to the basic simulation. Proper simulation improves median and percentage errors such as \gls{mae}, \gls{mape}, and \gls{smape}, while basic simulation obtains a slightly lower \gls{rmse}. This suggests that calibrated simulation helps the model under typical conditions but can slightly increase tail risk during extreme traffic situations.

\begin{table}[H]
\centering
\caption{Performance comparison of HINT against baseline models across three datasets (Lower is better).}
\label{tab:combined_metrics}
\resizebox{\textwidth}{!}{
\begin{tabular}{l|rrrr|rrrr|rrrr}
\toprule
& \multicolumn{4}{c|}{MOW} & \multicolumn{4}{c|}{UTD19-Torino} & \multicolumn{4}{c}{UTD19-Essen} \\
\cmidrule(lr){2-5} \cmidrule(lr){6-9} \cmidrule(lr){10-13}
Model & MAE & RMSE & MAPE(\%) & SMAPE(\%) & MAE & RMSE & MAPE(\%) & SMAPE(\%) & MAE & RMSE & MAPE(\%) & SMAPE(\%) \\
\midrule
KNN & 640.48 & 1023.96 & 717.43 & 121.38 & 238.67 & 378.02 & 114.47 & 0.6805 & 128.89 & 191.34 & 55.09 & 56.85 \\
OKriging & 448.85 & 891.14 & 417.52 & 93.40 & 226.44 & 342.56 & 108.67 & 62.90 & 113.29 & 156.86 & 69.22 & 46.64 \\
IGNNK & 529.18 & 963.05 & 603.92 & 108.69 & 232.68 & 351.42 & 142.36 & 66.21 & 123.80 & 177.20 & 109.38 & 61.51 \\
IAGCN & 480.17 & 996.47 & 457.64 & 99.87 & 224.01 & 350.40 & 112.97 & 63.52 & 112.88 & 164.59 & 65.63 & 48.07 \\
KITS & 317.19 & 758.69 & 210.4 & 71.62 & 213.10 & 327.44 & 154.75 & 64.56 & 97.94 & 148.45 & 58.68 & 42.10 \\
Ours - Basic simulation & 182.60 & \textbf{400.43} & 79.63 & 56.03 & \textbf{166.71} & \textbf{250.39} & \textbf{80.73} & \textbf{50.68} & \textbf{86.02} & \textbf{122.62} & \textbf{51.31} & \textbf{41.60} \\
Ours - Proper simulation & \textbf{158.66} & 416.74 & \textbf{68.62} & \textbf{47.33} & --- & --- & --- & --- & --- & --- & --- & --- \\

Ablation-Without Simulation & 210.18 & 436.65 & 130.62 & 59.92 & 175.04 & 267.17 & 82.86 & 52.58 & 110.03 & 158.26 & 58.05 & 51.78 \\
\bottomrule
\end{tabular}
}
\end{table}

\subsection{Comparison with Existing Inductive GNNs}
The proposed framework aims to reconcile two major challenges in modern traffic monitoring: (1) bridging the gap between widely available but "flow insensitive" speed data and (2) providing accurate flow estimates at locations lacking sensors. The results demonstrate that combining transductive speed modeling with inductive flow imputation significantly improves performance over purely inductive methods.

Recent work (e.g., IGNNK \citep{wu2021inductive}, INCREASE \citep{zheng2023increase}, KITS \citep{xu2023kits}) has highlighted the power of inductive GNNs for spatiotemporal kriging. However, most prior models are evaluated primarily on smoother variables like speed or environmental data (e.g., air quality), where nearby nodes exhibit more coherent patterns. In contrast, traffic flow can vary drastically between adjacent segments, especially on ramps and merges. By integrating transductive speed data and geospatial external features, our model overcomes abrupt disparities and heterophilous behavior of traffic flow.

\section{Conclusions}
This paper introduced an \textbf{INDU-TRANSDUCTIVE} training strategy to tackle the challenging problem of traffic flow imputation at unmeasured locations. By treating speed as a transductive variable that is broadly available from navigation apps and applying inductive GNN methods for flow, we combined the strengths of both paradigms. Additionally, we demonstrated that traffic simulations are useful traffic flow Kriging where even an uncalibrated simulation results can improve the data-driven model's accuracy. Our approach outperforms existing baselines and advanced GNN-based methods across multiple real-world datasets, showing especially strong performance on the MOW dataset. The more pronounced in accuracy improvement for the MOW dataset mainly stems from the fact the utilized simulation results for it are calibrated.

\noindent \textbf{Key findings include:}
\begin{itemize}
\item \textbf{Heterophily assumption:} Most \gls{gnn}s assume homophily which does not apply in traffic flow in diverse road networks our novel training strategy and the proposed model make heterophilous for both trends and scales.

\item \textbf{Synergy of Speed and Flow:} Leveraging transductive speed data bridges the gap between classical speed-flow diagrams and purely data-driven GNN approaches, yielding notable error reductions.

\item \textbf{Traffic simulation results and external features as input to the model:} During training the model sees flow on a randomly sampled subset of nodes and learns to reconstruct the rest, meanwhile, the hold-out nodes are never seen by the model. Because the traffic flow can differ vastly between nearby roads, the simulation can help the model determine the road types, expected traffic in those roads, etc.

\item \textbf{Robustness to Heterogeneity:} Driving distance based adjacency matrix, OSM-based feature similarity, and simulated traffic flows help address structural and functional differences (e.g., ramps vs. main roads).

\item \textbf{Scalability and Flexibility:} Our inductive approach scales to new sensor deployments, which is crucial for dynamic traffic networks.
\end{itemize}
While we report strong performance, the results also confirm the inherent complexity of flow estimation compared to speed imputation which is consistent with earlier observations by \citet{xu2023kits}.

\subsection{Limitations and Future Directions}
Although the proposed INDU-TRANSDUCTIVE framework achieves consistent gains, several limitations merit attention and open avenues for further work.

Our approach assumes network‐wide access to traffic speed and exploits it transductively. In the present study, however, speed is primarily obtained from loop detectors rather than \gls{fcd}. This mismatch may overestimate performance in deployments where \gls{fcd} speeds are noisier, irregularly sampled, or subject to penetration‐rate biases. A natural next step is to compile a combined dataset in which flow originates from loop detectors and speed from \gls{fcd}, with careful time alignment and harmonized aggregation. Such a resource would enable a more faithful evaluation of HINT under realistic sensing conditions and allow sensitivity analyses to \gls{fcd} coverage, sampling sparsity, and latency.

Results on UTD19-Essen indicate that removing simulation features degrades performance relative to KITS, suggesting that the local speed-flow coupling may be weaker and/or stronger homophily in this network. Future work could explore hybrid training schedules that borrow KITS’ incremental strategy within our INDU-TRANSDUCTIVE regime to better exploit any residual homophily where present.

Finally, broader validation across cities, seasons, and road classes is warranted. Cross‐city transfer would test the portability of the learned inductive components and the stability of the node‐wise calibration layer across network typologies.

\section*{Acknowledgements}
This work was partially funded by the OptiRoutS project (HBC.2022.0096). We would also like to thank the Flanders Department of Mobility and Public Works (MOW) for providing the Antwerp dataset and dynamic simulation results, and Transport \& Mobility Leuven (TML) for the dynamic simulation results.

\bibliographystyle{plainnat}
\bibliography{references}

\end{document}